\documentclass[letterpaper, 10pt, conference]{ieeeconf}
\IEEEoverridecommandlockouts
\usepackage[utf8]{inputenc}
\usepackage{float}
\usepackage{hyperref}
\usepackage{amssymb}
\usepackage{amsmath}
\usepackage{mathtools}
\usepackage{listings}
\usepackage{xcolor}

\makeatletter
\@ifundefined{labelindent}{}{\let\labelindent\relax}
\makeatother
\usepackage{enumitem}

\usepackage{graphicx}
\graphicspath{{figures/}}
\usepackage{mdframed}
\usepackage[version=4]{mhchem}
\usepackage{siunitx}
\usepackage{placeins}
\usepackage{booktabs}
\usepackage{longtable,tabularx}
\usepackage{url}
\usepackage{wrapfig}
\usepackage{hyperref}
\hypersetup{
  colorlinks=false,
  linkbordercolor={white},
}

\title{Hamilton--Jacobi Reachability for Spacecraft Collision Avoidance}

\author{
Larry Hui$^{1}$,
Jordan Kam$^{2}$,
William Su$^{2}$,
Jianshu Zhou$^{3}$, \textit{Member, IEEE}%
\thanks{This research is supported by the National University of Singapore under the NUS Start-up Grant (FY2026), corresponding to Jianshu Zhou (jianshuzhou@nus.edu.sg)}%
\thanks{$^{1}$Department of Mechanical Engineering, University of California, Berkeley, Berkeley, CA 94720, USA.}%
\thanks{$^{2}$Aerospace Engineering Program, University of California, Berkeley, Berkeley, CA 94720, USA.}%
\thanks{$^{3}$Department of Mechanical Engineering, National University of Singapore, Singapore 117575.}%
}

\begin{document}

\maketitle
\thispagestyle{empty}
\pagestyle{empty}

\begin{abstract}
This article presents a Hamilton--Jacobi (HJ) reachability framework for a two--satellite collision avoidance problem operating in the same circular orbit, where relative motion is modeled in the radial--tangential--normal (RTN) frame using planar Hill--Clohessy--Wiltshire (HCW) dynamics. We define the target state space as unsafe relative configurations in the orbit plane corresponding to minimum separation requirements consistent with Federal Communications Commission (FCC) orbital standards. The interaction between spacecraft is formulated as a zero--sum differential game, where Player 1 is the controlled satellite and Player 2 is modeled as a bounded adversarial disturbance with unknown intent. We present the HJ formulation and compute backward reachable sets that characterize relative states from which collision cannot be avoided under worst-case disturbances, while states outside this set admit provably collision-free trajectories. These reachable sets are integrated with supervisory hybrid control logic to determine when evasive maneuvers must be initiated, enabling mathematically grounded safety guarantees for scalability.
\end{abstract}

\section{Introduction}
Low-Earth orbit (LEO) has seen a rapid expansion of the operations of small spacecraft in the same orbit regime \cite{liou2008instability}. This boom in the use of satellites can be seen through constellations like SpaceX's Starlink network \cite{spacex_starlink}. While these constellations bring new opportunities for connectivity and Earth--observation, their scalability brings into question the sustainability of existing space traffic management (STM) architectures \cite{abdellatif2023caching, johnson2004space}. With future plans to increase the size of these constellations from $n=8000$ to $n=42000$ spacecraft, novel STM architectures to guarantee collision--avoidance could enable this vision \cite{spacex_starlink}. With such large constellations, traditional ground--based approaches to deconfliction in--orbit and strategic maneuvers planning is limited. Collision avoidance must be augmented with automated safety layers. STM today is highly centralized, with a human controller in the loop making these complex and often safety--critical decisions \cite{johnson2004space}. In cases where decisions must be made about a controlled spacecraft without insight of the other operator's intention, safe autonomy could drive down the uncertainty in planning maneuvers \cite{starek2015spacecraft}. Formal methods such as reachability analysis \cite{asarin2006recent} and differential game formulations \cite{achdou2013finite} offer powerful tools for characterizing states as potentially unsafe regions and synthesizing provably safe behaviors for collision avoidance applications.

\section{Related Work}

Hamilton-Jacobi (HJ) reachability is a formal methodology for guaranteeing safety in dynamical systems under bounded uncertainty \cite{bansal2017hamilton}. HJ reachability casts safety as a differential game between a controlled system and adversarial disturbances, and computes a backward reachable set (BRS) via a HJ-Isaacs partial differential equation whose viscosity solution characterizes all states that can lead to an unsafe condition under worst-case disturbances. This BRS approach has seen widespread adoption in air traffic management and collision avoidance research \cite{mitchell2005time, bayen2007aircraft}. Early work applied HJ BRS to detect potential ``loss of separation'' events between aircraft operating in shared airspace by formulating conflict detection as a dynamic game between standard control inputs and uncertainty in aircraft motion \cite{mitchell2003overapproximating}. Extensions of these ideas have also been applied to automated aerial refueling \cite{ding2008reachability} and other proximity operations, where reachable sets characterize safe relative configurations under bounded disturbances and limited control authority, and have been leveraged to inform supervisory decisions and hybrid maneuver execution \cite{ding2012methods}. By synthesizing switching conditions between pre-computed maneuver primitives, these methods provide mathematical guarantees that overcome the limitations of purely probabilistic screening approaches.

\begin{figure}[b]

    \centering
    \includegraphics[width=1\linewidth]{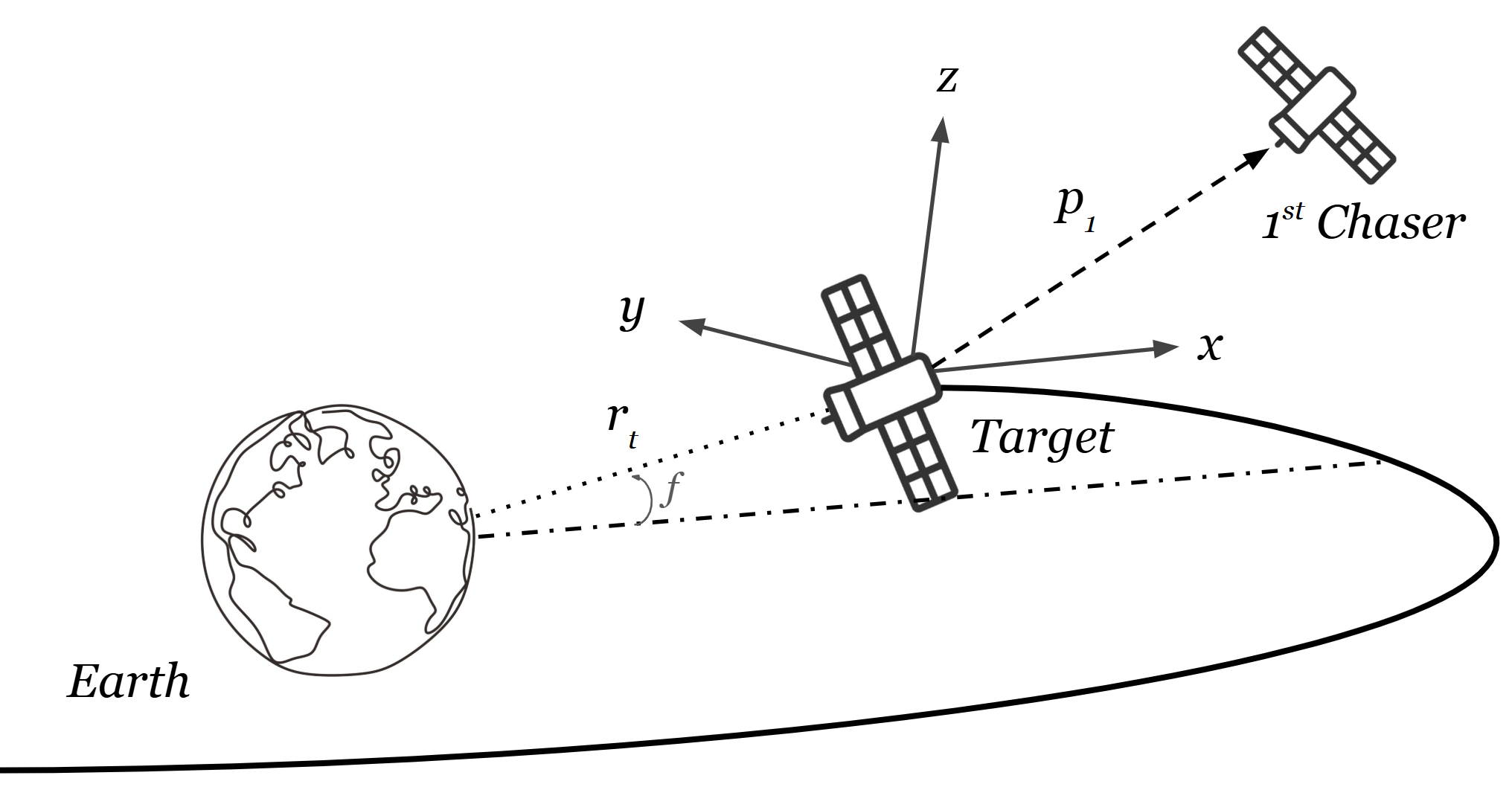}
    \caption{Relative-motion geometry in the RTN frame showing the target spacecraft and controlled chaser.}
    \label{fig:placeholder}
\end{figure}

For separation analysis, alternative stochastic reachability and Markov-chain based reachability techniques have been investigated; yet, deterministic HJ reachability is still unique in that it offers explicit worst-case safety margins under limited disturbances \cite{stochReach2019}. Applying these successes from aviation, satellites require different dynamics and constraints. In LEO, relative motion between spacecraft is commonly expressed in the radial–tangential–normal (RTN) frame and approximated locally by linear Hill–Clohessy–Wiltshire (HCW) dynamics. Unlike aircraft, spacecraft have limited control authority and propellant constraints, and maneuvers such as large altitude changes are often infeasible without impact on the mission. Several prior efforts have applied HJ reachability ideas to relative motion models and rendezvous/proximity operations, using backward reachability to characterize safe abort or contingency maneuvers under worst-case disturbances. \cite{kolemenFormation, 9938411, 7500704} Building on prior work, the present study formulates a two-satellite collision avoidance problem in the same circular orbit as a differential game in the HCW/RTN frame. 

\section{Methodology}

We consider a two-satellite collision avoidance scenario in which a controlled satellite (Player 1) operates in proximity to a secondary or uncooperative satellite (Player 2) within the same circular orbit. The relative motion between the two spacecraft is expressed in the RTN frame and approximated using planar HCW dynamics, which provide a locally linear representation of relative orbital motion. The state of the system is defined by the relative position and velocity vector $X = [x, y, \dot{x}, \dot{y}]^{\top}$, where $x$ denotes the tangential displacement and $y$ denotes the radial separation. Control inputs available to Player 1 are modeled as bounded accelerations $u \in U$, while uncertainty in the motion of Player 2 is represented as an adversarial disturbance $d \in D$. The collision avoidance problem is formulated as a zero-sum differential game in which Player 1 seeks to avoid unsafe relative configurations while Player 2 acts as a worst-case disturbance. We define a target set $\mathcal{T} \subset X$ representing unsafe orbital separations consistent with minimum separation standards, and compute the BRS that captures all initial states from which collision cannot be avoided despite optimal evasive control. States outside the BRS admit collision-free trajectories under admissible control actions, enabling the synthesis of formal safety guarantees and supervisory decision logic for maneuver execution.


\subsection{Satellite Model}


We will examine the interaction between a single controlled satellite (Player 1) and a secondary or uncooperative satellite (Player 2). We aim to study the continuous behavior of the two spacecraft in relative coordinates. Note, this is a valid simplification as in large constellations, de-centralized de-confliction architectures manage interactions at the individual vehicle level. To model this relative motion, we begin with the equations for the restricted 3-body problem.

\begin{equation}
\begin{aligned}
m\Bigl(&\frac{d^2\mathbf{r}}{dt^2}
+2\boldsymbol{\omega}\times\frac{d\mathbf{r}}{dt}
+\boldsymbol{\omega}\times(\boldsymbol{\omega}\times\mathbf{r})\Bigr) \\
&=
-\frac{Gmm_1}{R_1^3}\mathbf{R}_1
-\frac{Gmm_2}{R_2^3}\mathbf{R}_2
\end{aligned}
\end{equation}

where $m$ is the spacecraft mass, $\mathbf{r}$ is the absolute position vector, $\mathbf{R}_1$ and $\mathbf{R}_2$ are position vectors relative to the primary bodies, and $\boldsymbol{\omega}$ is the orbital angular velocity. Let the chase spacecraft and target spacecraft have masses $m$ and $m_1$, respectively. If $m$ and $m_1$ are treated as infinitesimal relative to the central body mass $m_2$, then $\mathbf{r}$ and $\mathbf{R}_2$ coincide. This allows the problem to be modeled as relative motion between a chase spacecraft and a target spacecraft orbiting a central body of mass $m_2$. We get the following equations of motion:

\begin{equation}
    \frac{d^2\mathbf{R}}{dt^2}+2\boldsymbol{\omega}\times \frac{d\mathbf{R}}{dt}+\boldsymbol{\omega}\times(\boldsymbol{\omega}\times (\mathbf{R}+\boldsymbol{r}_1))=-\frac{\omega^2r_1^3}{r^3}\textbf{r}
\end{equation}
The differential equation for the motion is initially non-linear due to the gravitational term. However, applying first order linearization yields the general vector differential equation for the relative motion of the chase spacecraft:
\begin{equation}
    \frac{d^2 \mathbf{R}}{dt^2}+2\boldsymbol{\omega}\times \frac{d\mathbf{R}}{dt}=-\omega^2 \mathbf{R}+3\omega^2 (i_\xi \cdot \mathbf{R})i_\xi
\end{equation}
where the relative position vector $\mathbf{R}=xi_\theta + y i_r -zi_z$. $x$ is the tangential direction, $y$ is the radial direction, and $z$ is normal to the orbital plane. This yields three simultaneous second-order, linear, constant-coefficient differential equations. Because the two satellites operate in the same circular orbit and our primary concern is in-plane collision risks, we restrict our focus to a 2D planar solution. The out-of-plane $z$-axis motion is decoupled and omitted, leaving the unforced planar HCW equations:
\begin{equation}
    \begin{dcases}
\frac{d^2x}{dt^2}+2\omega \frac{dy}{dt}=0\\
\frac{d^2y}{dt^2}-2\omega \frac{dx}{dt}-3\omega^2y=0
\end{dcases}
\end{equation}

The planar HCW model is appropriate for illustrating the proposed scenario in a locally linear same orbit encounter setting, but it omits several effects relevant to operational deployment. In particular, out-of-plane relative motion, secular perturbations such as $J_2$, and nonlinear orbital effects are not captured in the present analysis. As a result, the computed reachable sets should be interpreted as formal guarantees with respect to the adopted planar linearized model. Extending the framework to 3D relative motion is a natural next step for practical deployment.

In order to formulate this collision-avoidance scenario as a two-player differential game, we convert the HCW dynamics into a state-space formulation. Let us define the state vector $X\triangleq [x,y,\dot x,\dot y]^\top\in \mathbb R^4$. We can then introduce a control input vector $u\triangleq[u_x,u_y]^\top\in \mathcal U$ for Player 1 and a bounded disturbance vector $d\triangleq[d_x,d_y]^\top\in \mathcal D$ for Player 2. Then, the continuous-time state evolution equation $\dot X=AX+B_uu+B_dd$ is:

\begin{equation}
    \dot X=f(X,u,d)= \begin{bmatrix}
    \dot x\\\dot y\\-2\omega \dot y + u_x + d_x\\2\omega \dot x + 3\omega^2 y + u_y + d_y
\end{bmatrix}
\end{equation}

Because our safety criteria (and the target state space for collision) depend entirely on the physical separation distance mandated by FCC guidelines, our system output only needs to observe the relative positional states. We define the output equation $Y=CX+Du$:
\begin{equation}
    Y=\begin{bmatrix}
    1 & 0 & 0 & 0\\0 & 1 & 0 & 0
\end{bmatrix}\begin{bmatrix}
    x\\y\\\dot x \\ \dot y
\end{bmatrix}+\textbf{0}\cdot u=\begin{bmatrix}
    x\\y
\end{bmatrix}
\end{equation}

Here, $x$ and $y$ denote the tangential and radial relative positions, $\dot{x}$ and $\dot{y}$ the corresponding relative velocities, $u_x$ and $u_y$ the control accelerations of Player~1, $d_x$ and $d_y$ the bounded disturbance accelerations of Player~2, and $\omega$ the angular velocity of the circular orbit, with $\omega^2 = Gm_2/r_1^3$.

\subsection{Hybrid System Model}

Given the continuous trajectories of the satellite in-orbit, and distinct control laws for different operational phases, a hybrid system model is used to represent the satellite's decision-making process \cite{tomlin2000game}. We model the decentralized collision-avoidance process as a sequence of transitions between a finite number of discrete states. In each state, the relative motion of the satellites evolves according to the planar HCW dynamics, but subject to any control inputs and objectives. In the specific case of decentralized satellite collision avoidance, we introduce a hybrid automaton that governs Player 1. In this model, we define the states as the set of operational modes $\mathcal Q=\{q_\text{nom}, q_{\text{eva}}, q_{\text{ret}}\}$, where $q_\text{nom}$ is the standard station-keeping or orbiting, $q_\text{eva}$ is the execution of an active collision-avoidance maneuver, and $q_\text{ret}$ is the trajectory recovery phase. 

The discrete transitions in our decentralized architecture are autonomous and governed by state-dependent guard conditions. Let the continuous state satisfy $X \in \mathcal X \subset \mathbb R^4$. For Player~1, the hybrid automaton switches modes according to the relative state with respect to the BRS and recovery regions. We define the mode-transition guards as:

\begin{equation}
\label{eq:hybrid_guards}
\begin{aligned}
q_\text{nom} \rightarrow q_\text{eva} &:\; X \in \partial \mathcal G(\tau),\\
q_\text{eva} \rightarrow q_\text{ret} &:\; X \in \mathcal X_{\text{safe}},\\
q_\text{ret} \rightarrow q_\text{nom} &:\; X \in \mathcal G_0^{\text{rec}},\;
|\dot x| \le \varepsilon_x,\;
|\dot y| \le \varepsilon_y.
\end{aligned}
\end{equation}

where $\mathcal G(\tau)$ denotes the unsafe backward reachable set over horizon $\tau$, $\mathcal X_{\text{safe}} \subset \mathcal X \setminus \mathcal G(\tau)$ denotes a recovered safe region with sufficient separation margin, and $\mathcal G_0^{\text{rec}}$ denotes the recovery target set. Thus, the boundary of the BRS serves as the critical switching surface that determines when nominal operation must be abandoned in favor of evasive control, while the latter two guards certify when the spacecraft has recovered sufficient clearance to begin stabilization and eventual return to nominal operation. The continuous inputs consist of the control effort of Player~1, $u\in\mathcal U$, and the bounded external disturbance representing Player~2, $d\in\mathcal D$, with continuous state evolution governed by Eq.~(5). With this hybrid system formalism now established, we can define the formal verification and reachability problems that we wish to solve for safety-critical space environments. First, we define the target set $\mathcal T \subset \mathcal X$ as the set of unsafe relative positions in the orbit plane that violate the minimum orbit separation standard mandated by the FCC.

Our primary objective is to find the set of states from which Player 2 could force Player 1 into a collision, despite Player 1's best evasive actions. Because Player 1 has no information on the intent of Player 2, we can frame this as a zero-sum differential game. To formally define this, let $\mathcal V=\mathcal U\cup \mathcal D$ be the set of continuous input variables, where $\mathcal U$ is the set of continuous control inputs and $\mathcal D$ is the set of continuous disturbance inputs. The continuous evolution in each mode is described by a vector field $f:\mathcal Q\times \mathcal X\times \mathcal V\rightarrow \mathcal X$, where $q\in \mathcal Q$ indexes the active discrete mode. This vector field $f$ is assumed to be globally Lipschitz in $X$ (for a fixed $q\in \mathcal Q$) and continuous in $\mathcal V$. Also note that discrete transitions between modes are described by a reset relation: $R:\mathcal Q\times \mathcal X\times \Sigma \times \mathcal V\rightarrow 2^{\mathcal Q\times \mathcal X}$. 

Now, we can define a time-dependent BRS, denoted $\mathcal G(\tau)\subset \mathcal X$, for a finite time horizon $\tau \leq 0$. For the evasive mode $q_{\text{eva}}$, the problem becomes finding $\mathcal G(\tau)$ such that if $X(0)\in \mathcal G(\tau)$, then under the mode-dependent continuous vector field $f(q,X,u,d)$, the system state will eventually enter the target set $X(\tau)\in \mathcal T$, optimized over the worst-case disturbance $d\in \mathcal D$ against the optimal evasive control $u\in \mathcal U$. This formulation yields the unsafe set for the system. Consequently, robust formal verification is achieved: if the hybrid automaton ensures that Player~1 initiates the transition to $q_\text{eva}$ before the state reaches the boundary of $\mathcal G(\tau)$, the system admits mathematically guaranteed collision-free trajectories for standard satellite operations. 

\subsection{Control Law Design}

The feedback control laws for the controlled satellite (Player 1) to perform any standard orbital actions and state transitions are applied via the continuous thrust inputs $u_x$ and $u_y$, which represent the applied tangential and radial accelerations, respectively. For simplicity, we utilize Proportional-Derivative (PD) control to drive Player 1 to our desired target states. For transitions between stationary orbital slots, the equations of the feedback laws are given by:
\begin{align}
    u_x&=-k_{p,x}(x-x_f)-k_{d,x}(\dot x - \dot x_f)\\
    u_y&=-k_{p,y}(y-y_f)-k_{d,y}(\dot y - \dot y_f)
\end{align}
where $k_{p,x}, \,k_{p,y}$ are the proportional controller gains, and $k_{d,x}, \,k_{d,y}$ are the derivative controller gains. $(x_f,y_f)$ and $(\dot x_f,\dot y_f)$ are the desired final relative position and velocity of Player 1 in the RTN frame, respectively. Note that we omit an integral control term due to space's lacks constant external disturbances that result in steady-state errors. To model saturation in the propulsion system, we restrict the above control inputs to lie within the satellite's maximum thrust capabilities: $u_x\in [-u_{\max}, u_{\max}]$ and $u_y \in [-u_{\max}, u_{\max}]$

At any point during standard operations, evasive maneuvers may need to be performed in an attempt to avoid an imminent collision between Player 1 and the uncooperative Player 2. This could occur if Player 1 deviates from its standard trajectory due to external orbital perturbations, or if bounded disturbances from Player 2 force the system state toward the boundary of the BRS. When a safety violation is imminent, the hybrid automaton will transition from $q_{\text{nom}}$ into the evasive state $q_{\text{eva}}$, the system abandons the nominal target and we can define four different escape modes based on relative locations within the planar RTN frame. The modes map directly to standard V-bar (velocity vector) and R-bar (radial vector) escape maneuvers:

\vspace*{-.05cm}
\begin{enumerate}
    \item \textbf{Mode 1}: Player 1 attempts to move to a higher altitude (positive radial) which naturally causes it to drift backward relative to Player 2 (negative tangential).
    \item \textbf{Mode 2}: Player 1 attempts to move to a lower altitude (negative radial) which naturally causes it to accelerate forward relative to Player 2 (positive tangential).
    \item \textbf{Mode 3}: Player 1 executes a pure tangential braking maneuver to increase separation distance behind Player 2 without altering its radial altitude.
    \item \textbf{Mode 4}: Player 1 executes a pure tangential acceleration maneuver to get ahead of Player 2 on the same altitude.
\end{enumerate}

\begin{figure}
[t]
    \centering
    \includegraphics[width=1\linewidth]{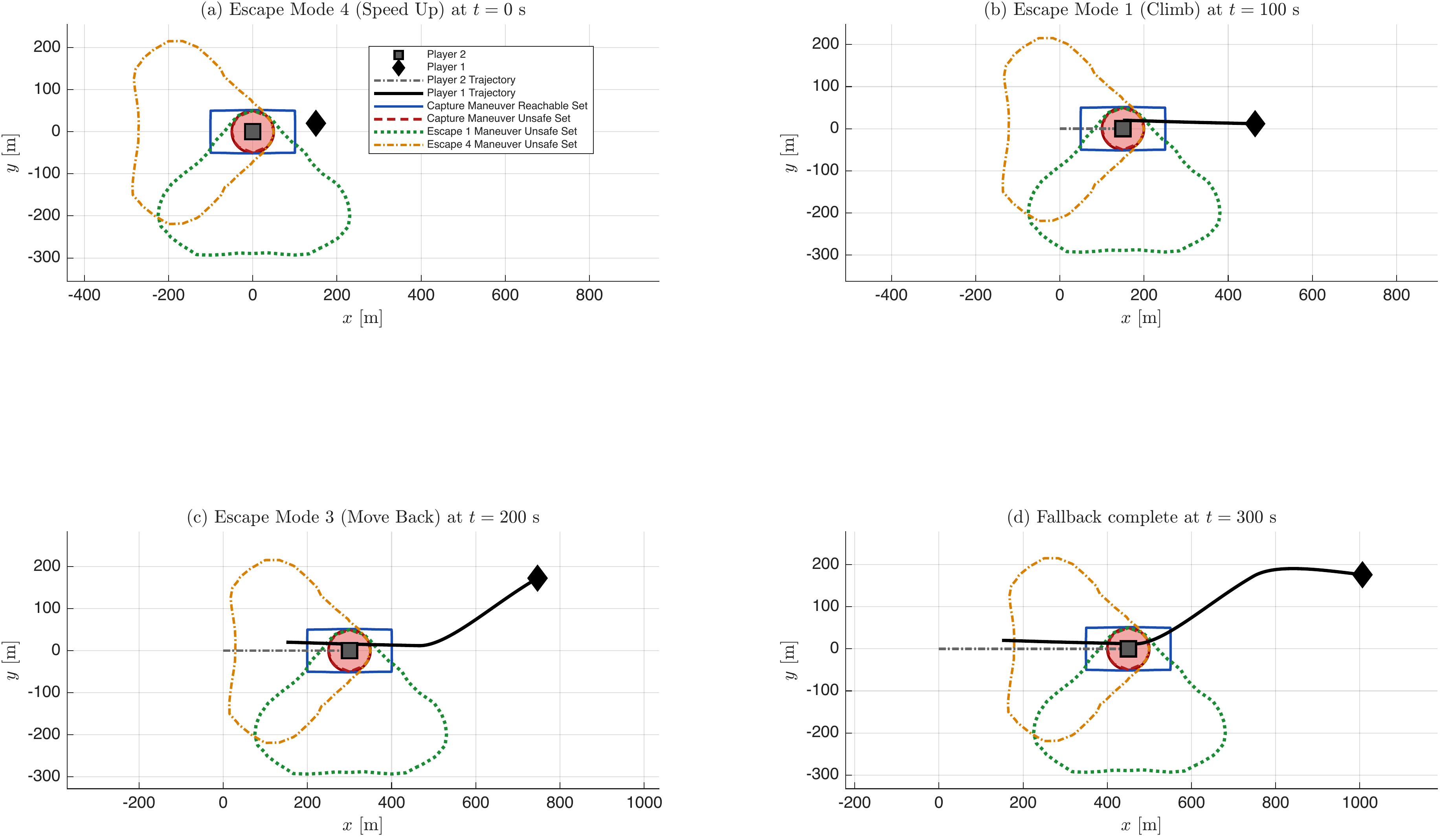}
    \caption{Escape maneuvers for the satellite fallback scenario.}
\end{figure}

\section{Results}

We first review the general HJ method allowing for input uncertainty, and then adapt it to our specific system dynamics.\cite{bansal2017hamilton} Recall that we defined the continuous relative state of the system in the RTN frame to be $X=[x,y,\dot x, \dot y]^\top\in \mathbb R^4$. Then the continuous evolution of the satellites is governed by the following state dynamics $\dot X=f(X,u,d)$ where $u\in \mathcal U$ represents the continuous evasive action control inputs of Player 1, and $d\in \mathcal D$ represents the bounded disturbance inputs modeling the unknown intent of Player 2. We can define the target set implicitly as the sublevel set of a continuous scalar function of the states, $\phi_0:\mathbb R^4\rightarrow \mathbb R$. In our case, the target set, denoted $\mathcal G_0$, represents the unsafe regions in the orbit plane that would likely result in a collision or violate the FCC orbit separation standard. We define this target set such that a state $X$ is within the unsafe collision zone if and only if the level set function $\phi_0(X)$ is less than or equal to zero, mathematically:
\begin{equation}
     \mathcal{G}_0 = \{ X \in \mathbb{R}^4 \mid \phi_0(X) \leq 0 \}
\end{equation}
Similarly, one can define the set of initial relative states from which a collision can be forced after time $\tau$ as the sublevel set of the time-dependent level set function $\phi : \mathbb{R}^4 \times \mathbb{R} \rightarrow \mathbb{R}$. Let this BRS be $\mathcal{G}(\tau)$, then:
\begin{equation}
\mathcal{G}(\tau) = \{ X \in \mathbb{R}^4 \mid \phi(X,-\tau) \leq 0 \}
\end{equation}
Recall that we formulated this scenario as a zero-sum differential game over a finite time horizon $T$. Let $\xi_{X,T}^{u,d}(t)$ be the path that a system takes starting from an initial state $X$ at time $t=0$, evolving under some control input $u$ and bounded disturbance $d$. Since we consider propellant expenditure secondary during an evasive maneuver, we will not employ a running cost as we only care about whether the system actually enters the target set $\mathcal G_0$ at any point over $T$ or not. However, future work could incorporate a running cost $L(X,u)$ within the Hamiltonian to explicitly capture fuel-optimal evasive behavior. Thus, we evaluate the minimum value of $\phi_0(X)$ along the path. The optimal value function $\phi(X,T)$ is given by
\begin{equation}
     \phi(X,T)=\inf_{u\in \mathcal U} \sup_{d\in\mathcal D} \min_{t\in[0,T]}\phi_0(\xi_{X,T}^{u,d}(t))
\end{equation}
The set of states where $V(X,t)\leq 0$ defines the BRS. It represents the set of initial relative states from which Player 2 can force a collision despite Player 1's optimal evasive actions. If all inputs within spaces $\mathcal U$ and $\mathcal D$ are bounded and the system behaviour $f(X,u,d)$ satisfies continuity constraints, the evolution of this reachable set boundary is described by the HJI PDE. Since we only evaluate the terminal/target conditions without a running cost, we can formulate the optimal control Hamiltonian $\mathcal H$ around the system dynamics and the spatial gradient of the value function, $\nabla \phi$:
\begin{equation}
\begin{aligned}
    \mathcal H(X,\nabla \phi)=\min_{u\in \mathcal U}\max _{d\in\mathcal D} \langle f(X,u,d), \nabla \phi\rangle\\ 
    =-\max _{d\in\mathcal D}\min_{u\in \mathcal U}\nabla \phi^\top f(X,u,d)
\end{aligned}
\end{equation}
Then, we can write the standard HJI PDE update equation as
\begin{equation}
\begin{aligned}
\frac{\partial \phi}{\partial t} + \mathcal H(X,\nabla \phi) &= 0, \\
\phi(X,0) &= \phi_0(X), \\
-\frac{\partial \phi}{\partial t} &\leftarrow \mathcal H\left(X,\frac{\partial \phi}{\partial X}\right).
\end{aligned}
\end{equation}
The HJI PDE computes states that end up exactly in $\mathcal G_0$ at $T$. However, to guarantee safety, we must compute the Backward Reachable Tube (BRT), the set of all states that could enter $\mathcal G_0$ at any time during the horizon. To achieve this, the value function must not be allowed to increase once a trajectory has breached the target set $\phi_0(X) \leq 0$.To compute this, we use the following variational inequality,
\begin{figure}
[t]
    \centering
    \includegraphics[width=.95\linewidth]{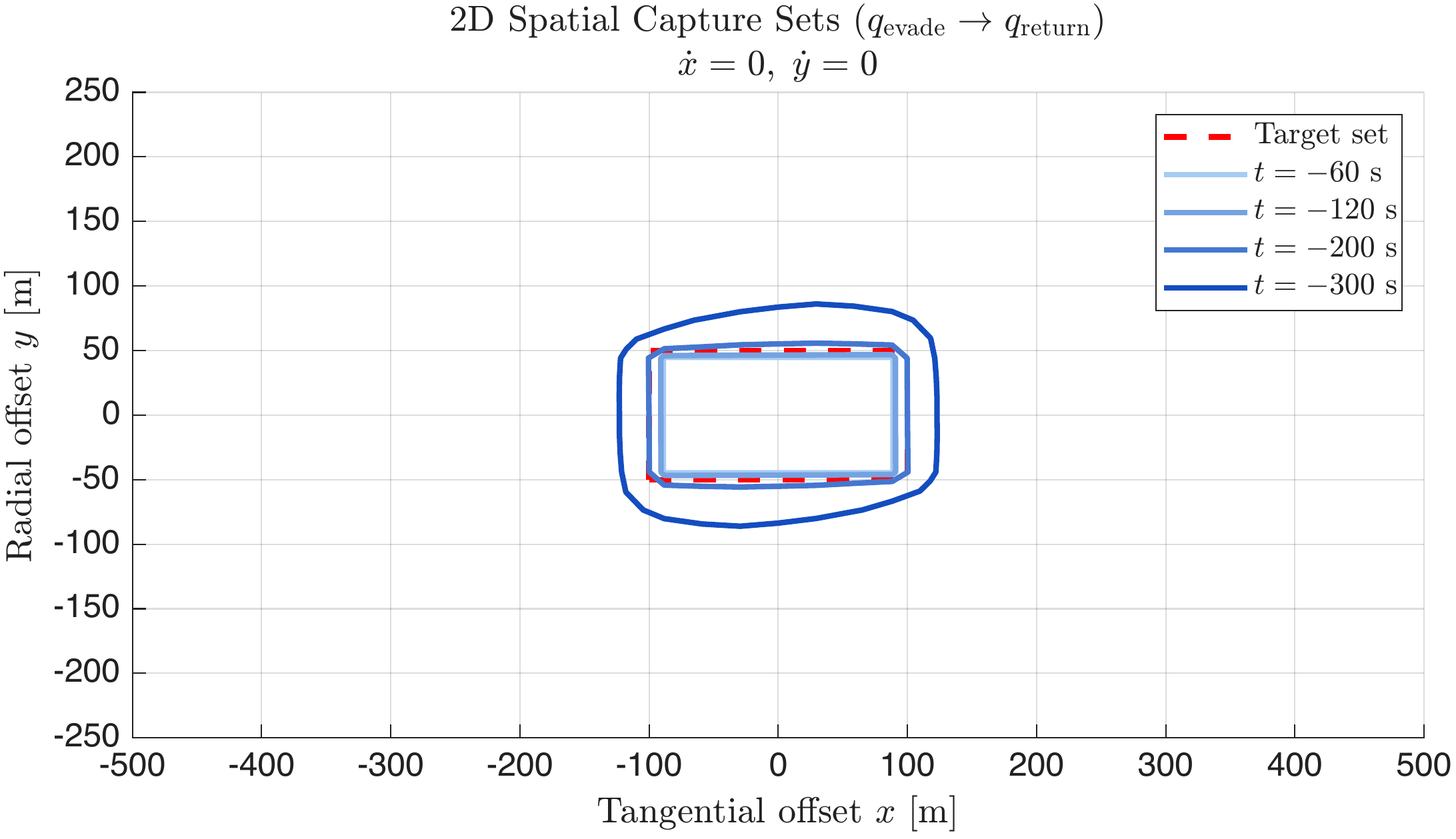}
    \caption{2D Capture Set for Recovery $q_{\text{eva}}$ to $q_{\text{ret}}$ slice at 3 radial velocities.}
    \label{fig:placeholder}
\end{figure}

\begin{equation}
    \min \left( \frac{\partial \phi}{\partial t} + \mathcal{H}(X, \nabla \phi), \ \phi_0(X) - \phi(X, t) \right) = 0
\end{equation}
For general system dynamics, the solution to this HJI PDE and variational inequality is difficult to compute analytically. Several high-resolution numerical approximation schemes exist to compute the level set function. Computations of these reachable sets are typically performed using Level Set Methods, wherein the continuous state space is divided into a finite number of grid cells, and each cell is assigned a numerical value of the level set function during the backward integration.


We note that for the hybrid system automaton defined in Section III-C, the set of disturbance inputs is empty. To generate the reachable sets for a particular feedback control law $u = K(X)$ as defined in Section IV-D, one would use the system behavior $\dot X = f(X,K(X))$, which is input free. With this formulation of the Hamiltonian, we are now able to generate the reachable sets for each of the formation transitions and evasive maneuvers with fixed control inputs. For capture sets, the target set is chosen to be a closed set of states around a desired waypoint. The general form of such a target set is given below:
\begin{equation}
    \mathcal G_0 = \begin{dcases}
    x\in [x_{\min}, x_{\max}]\\
    y\in [y_{\min}, y_{\max}]\\
    \dot x \in [\dot x_{\min}, \dot x_{\max}]\\
    \dot y \in [\dot y_{\min}, \dot y_{\max}]
\end{dcases}
\end{equation}
Conversely, for the collision-avoidance analysis, the target set for the unsafe BRS is chosen to be a hard physical boundary around Player 2. This represents the minimum safe separation distance mandated by the FCC, $d_{FCC}$. Precisely, this unsafe target set (representing the immediate collision zone) is defined purely by the spatial relative coordinates:

\begin{equation}
    \mathcal G_0=\{X\in \mathbb R^4 | x^2+y^2 \leq d_{FCC}^2\}
\end{equation}

We note, however, that the true strength of this HJ formulation is its ability to natively handle uncertainty in the uncooperative satellite's (Player 2) actions. By modeling Player 2's potential thrust or unmodeled orbital drift as a bounded disturbance, we add strict mathematical robustness to the generated safe and unsafe sets. Specifically, rather than assuming Player 2 follows a constant, predictable Keplerian trajectory, we allow its relative acceleration to vary dynamically within deterministic bounds. In the framework of the HJ equations, this worst-case disturbance input space for Player 2 is formally defined as:

\begin{equation}
    \mathcal{D} = \{ d \in \mathbb{R}^2 \mid d_x \in [d_{x,\min}, d_{x,\max}], \ d_y \in [d_{y,\min}, d_{y,\max}] \}
\end{equation}

The bounded disturbance set $\mathcal{D}$ yields worst-case safety guarantees, but is inherently conservative, as it allows all admissible disturbance realizations regardless of likelihood. As a result, the computed BRS may overapproximate practically unsafe states. Less conservative formulations could enlarge the usable safe set at the cost of reduced robustness. By solving the zero-sum differential game over this disturbance space $\mathcal{D}$, the resulting BRS guarantees that Player 1 can avoid $\mathcal{G}_0$ regardless of any allowable maneuver Player 2 attempts to execute. Analytical solutions to the HJI PDE are generally intractable. Therefore, we rely on high-resolution numerical approximation schemes to evaluate the time-varying level set function. We present computations of the BRS as performed using a modified Level Set Methods Toolbox \cite{mitchell2008flexible}, based upon the level set theory described extensively in \cite{mitchell2002application}. Within this numerical framework, the continuous 4D state space ($X \in \mathbb{R}^4$) of the planar HCW model is discretized into a finite multidimensional grid. Each discrete grid node is assigned a scalar numerical value representing the value function. Because the computational complexity of grid-based level set methods scales exponentially with the state-space dimension, the number of grid cells must be chosen carefully to balance the fidelity of the mathematical safety guarantees with the computational feasibility of the 4D simulation.

For larger constellations, several extensions are possible. First, structure in the dynamics may be exploited through decomposition or pairwise safety approximations, so that local relative-motion games are solved only for nearby vehicles rather than over a joint multi-agent state. Second, adaptive or nonuniform gridding could concentrate resolution near the target set and BRS boundary, where accuracy is most safety-critical. Third, surrogate approximations, including learned value-function representations, may accelerate online evaluation while retaining offline-computed safety structure. These directions are not pursued in this work, but they provide a practical path toward scaling the framework beyond the present 4D planar setting.

\begin{figure}[b]
    \centering
    \includegraphics[width=1\linewidth]{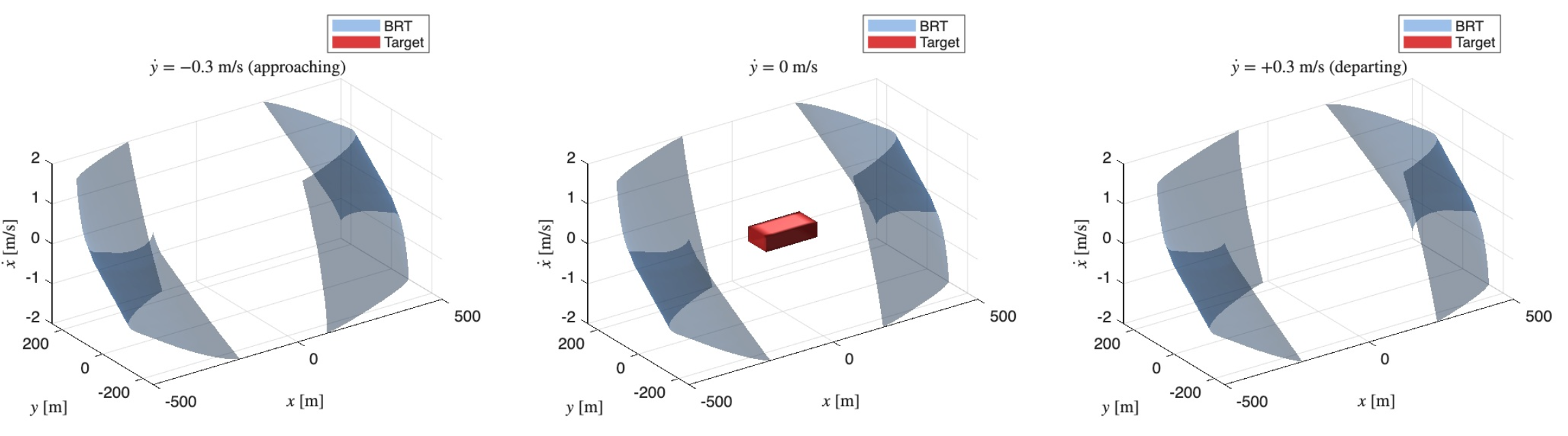}
    \caption{3D Capture Set for Recovery $q_{\text{eva}}$ to $q_{\text{ret}}$ slice at 3 radial velocities.}
\end{figure}

\subsection{Capture Sets Computation and Simulation}

To validate the mathematically guaranteed safe and unsafe subsets, we constructed full-state orbital simulations of various nominal and evasive operational scenarios. While the HJ BRS were computed numerically using Level Set Methods. For the simulation results shown in this section, we set the mean motion of the reference LEO circular orbit to 0.0011 rad/s (corresponding to an altitude of roughly 500 km). Furthermore, we assume the continuous control inputs (Player 1's thrust accelerations) have symmetric saturation limits of $[-0.1, 0.1]$m/s$^2$, and the uncooperative bounded disturbances of Player 2 are strictly less than Player 1's control authority, capped at $[-0.05, 0.05]$ m/s$^2$. This bound is selected as a conservative surrogate for uncooperative relative actuation, such as limited rogue thrusting, while remaining below the available control authority of Player~1. The PD control law parameter values used for the simulation are summarized in Table I.

\begin{table}[t]
\centering
\caption{Controller Gain Parameters}
\label{tab:gains}
\footnotesize
\setlength{\tabcolsep}{3.5pt}   
\renewcommand{\arraystretch}{1.05}

\begin{tabular}{lcccc}
\toprule
\textbf{Parameter} & \textbf{Symbol} & \textbf{Nominal/Recovery} & \textbf{Evasive} & \textbf{Units} \\
\midrule
Tangential P Gain & $k_{p,x}$ & $1.0\!\times\!10^{-4}$ & $5.0\!\times\!10^{-4}$ & $\mathrm{s^{-2}}$ \\
Radial P Gain     & $k_{p,y}$ & $1.0\!\times\!10^{-4}$ & $5.0\!\times\!10^{-4}$ & $\mathrm{s^{-2}}$ \\
Tangential D Gain & $k_{d,x}$ & $2.0\!\times\!10^{-2}$ & $4.4\!\times\!10^{-2}$ & $\mathrm{s^{-1}}$ \\
Radial D Gain     & $k_{d,y}$ & $2.0\!\times\!10^{-2}$ & $4.4\!\times\!10^{-2}$ & $\mathrm{s^{-1}}$ \\
\bottomrule
\end{tabular}
\end{table}

In terms of the PD control laws, the desired spatial location in the planar RTN frame is given by the target parameters $x_f$ and $y_f$. Since we require the controlled satellite to end its recovery maneuver fully stabilized relative to the reference slot, the desired final relative velocities ($\dot{x}_f, \dot{y}_f$) are zero.For example, the target recovery set for transitioning back to nominal station-keeping ($q_{\text{ret}} \rightarrow q_{\text{nom}}$) at a safe tangential standoff distance of 1000 m can be defined as the following four-dimensional bounding box:
\begin{equation}
    \mathcal G_0 = \begin{dcases}
    x\in [950, 1050]\\
    y\in [-25, 25]\\
    \dot x \in [-0.01, 0.01]\\
    \dot y \in [-0.01, 0.01]
\end{dcases}
\end{equation}


\section{Conclusion}

This article presents a HJ reachability framework for decentralized spacecraft collision avoidance that integrates formal safety verification with hybrid supervisory control logic. By modeling relative motion in the RTN frame using planar HCW dynamics and formulating spacecraft interactions as a zero-sum differential game under bounded disturbances, BRS are computed to rigorously characterize unsafe relative configurations and define the set of states from which collision cannot be avoided, while states outside this boundary admit provably collision-free trajectories. These reachable sets are incorporated within a hybrid automaton governing nominal operation, evasive maneuvers, and recovery phases, enabling mathematically grounded supervisory decisions for collision avoidance initiation and ensuring safe maneuver execution under worst-case uncertainty in the behavior of an uncooperative spacecraft. Numerical simulations performed using level set methods validate the structure of the safe and unsafe regions under realistic thrust constraints and disturbance bounds.

Future work will extend the framework to full 3D relative motion to incorporate out-of-plane dynamics, more realistic orbital perturbations and nonlinear orbital effects. Additional work will investigate scalable approximations and decomposition techniques to mitigate the computational burden associated with high-dimensional reachable set computation. Further studies would emphasize less conservative uncertainty descriptions, such as probabilistic or adaptive disturbance models.

\bibliographystyle{IEEEtran}
\bibliography{bib/references}

\end{document}